\documentclass[preprint,review,12pt]{elsarticle}

\usepackage{amssymb}
\usepackage{amsmath}
\usepackage{color}
\usepackage{multirow}
\usepackage{enumitem}
\usepackage{float}
\usepackage{booktabs}
\usepackage{subcaption}
\usepackage{graphicx}
\usepackage{epstopdf}

\journal{Pattern Recognition}

\begin{document}

\begin{frontmatter}


%Analytic Decision-Space Channel Gating: A Nonparametric, Explainable Module for Fine-Grained Feature Decoupling in YOLO
%Decision-Space-Driven Dynamic Gating for Fine-Grained Feature Decoupling in YOLO
%\title{Analytic Decision-Space Channel Gating: A Nonparametric, Explainable Module for Fine-Grained Feature Decoupling in YOLO}

\title{YOLO-DS: Fine-Grained Feature Decoupling via Dual-Statistic Synergy Operator for Object Detection}


\author[label1,label7,label8]{Lin Huang}
\ead{h72001346@163.com}

\author[label2]{Yujuan Tan  \corref{cor1}}
\cortext[cor1]{Corresponding Author}
\ead{tanyujuan@nudt.edu.cn}

\author[label3]{Weisheng Li}
\ead{liws@cqupt.edu.cn}

\author[label7,label8]{Shitai Shan}
\ead{shansht@inspur.com}

\author[label4]{Liu Liu}
\ead{244319450@qq.com}

\author[label5]{Bo Liu}
\ead{liubo@fawer.com.cn}

\author[label6]{Linlin Shen}
\ead{llshen@szu.edu.cn}

\author[label1]{Jing Yu}
\ead{20201401010@cqu.edu.cn}

\author[label1]{Yue Niu}
\ead{yuen@cqu.edu.cn}

\affiliation[label1]{organization={Chongqing University},%Department and Organization
            addressline={No.55, University Town South Road, Gaoxin District}, 
            city={Chongqing},
            postcode={401331},
            country={China}}

\affiliation[label2]{organization={National University of Defense Technology},%Department and Organization
            addressline={No.1, Fuyuan Road, Kaifu District}, 
            city={Changsha},
            postcode={410073},
            country={China}}

\affiliation[label3]{organization={Chongqing University of Posts and Telecommunications},%Department and Organization
            addressline={No.2, Chongwen Road, Nan'an District}, 
            city={Chongqing},
            postcode={400065},
            country={China}}

\affiliation[label4]{organization={China Academy of Information and Communications Technology},%Department and Organization
            addressline={No.52, Huayuan North Road, Haidian District}, 
            city={Beijing},
            postcode={100191},
            country={China}} 

\affiliation[label5]{organization={Fawer Automotive Components Co., Ltd},%Department and Organization
            addressline={No.777, Dongfeng South Street, Changchun Automobile Economic and Technological Development Zone}, 
            city={Changchun},
            postcode={130062},
            country={China}}                     
 
\affiliation[label6]{organization={Shenzhen University},%Department and Organization
            addressline={No.3688 Nanshan Avenue, Nanshan District}, 
            city={Shenzhen},
            postcode={518061},
            country={China}}
            
\affiliation[label7]{organization={Inspur Yunzhou Industrial Internet Co., Ltd},%Department and Organization
            addressline={No.1036 Langchao Road, Lixia District}, 
            city={Jinan},
            state={Shandong},
            postcode={250101},
            country={China}}

\affiliation[label8]{organization={Guoqi Zhimo (Chongqing) Technology Co., Ltd.},%Department and Organization
            addressline={5th Floor, Building B15, Xiantao Data Valley, Yubei District}, 
            city={Chongqing},
            postcode={401122},
            country={China}}

%% Abstract
\begin{abstract}

\iffalse
﻿﻿One-stage object detection is valued for its balanced accuracy-efficiency trade-off, with the YOLO series leading in both performance and deployment. However, existing YOLO detectors lack explicit modeling of heterogeneous object responses within shared feature channels, limiting further performance gains. To address this, we propose YOLO-DS, a framework centered on a Dual-Statistic Synergy Operator (DSO) that decouples object features by modeling channel mean and peak-to-mean difference. We also introduce a Dual-Statistic Synergy Gating (DSG) module for adaptive channel-wise feature selection and a Multi-Path Segmented Gating (MSG) module for depth-wise feature weighting. On MS-COCO, YOLO-DS consistently outperforms YOLOv8 across five scales (N, S, M, L, X) with AP gains of 1.4-1.7\% and minimal latency increase. Visualization, ablation, and comparative studies validate the approach, demonstrating effective discrimination of heterogeneous objects with high efficiency.
\fi

One-stage object detection, particularly the YOLO series, strikes a favorable balance between accuracy and efficiency. However, existing YOLO detectors lack explicit modeling of heterogeneous object responses within shared feature channels, which limits further performance gains. To address this, we propose YOLO-DS, a framework built around a novel Dual-Statistic Synergy Operator (DSO). The DSO decouples object features by jointly modeling the channel-wise mean and the peak-to-mean difference. Building upon the DSO, we design two lightweight gating modules: the Dual-Statistic Synergy Gating (DSG) module for adaptive channel-wise feature selection, and the Multi-Path Segmented Gating (MSG) module for depth-wise feature weighting. On the MS-COCO benchmark, YOLO-DS consistently outperforms YOLOv8 across five model scales (N, S, M, L, X), achieving AP gains of 1.1\% to 1.7\% with only a minimal increase in inference latency. Extensive visualization, ablation, and comparative studies validate the effectiveness of our approach, demonstrating its superior capability in discriminating heterogeneous objects with high efficiency.
﻿
\end{abstract}



%% Keywords

\end{frontmatter}



\section{Introduction}

One-stage object detection\cite{yolo1,yolo2,yolo3,yolo4,syolo4,yolo5,yolopp,yolopp2,yolof,yolox,yolo7,yolo8,yolo9,yolo10,yolo11,focal,efficientdet,asff,ssd,pr1,pr2,pr3,pr4} frameworks, especially the YOLO series, are renowned for their optimal trade-off between accuracy and speed, enabling widespread real-time applications. Nonetheless, a fundamental limitation persists: these detectors operate on a shared channel modeling scheme, failing to explicitly distinguish the distinct response patterns of heterogeneous objects (e.g., small vs. large, foreground vs. background). This homogenized representation triggers channel-wise competition and impedes dynamic attention allocation, ultimately capping detection performance.

\begin{figure}[H]
\centering
    \begin{minipage}[b]{0.48\linewidth}
        \centering
        \includegraphics[scale=0.31]{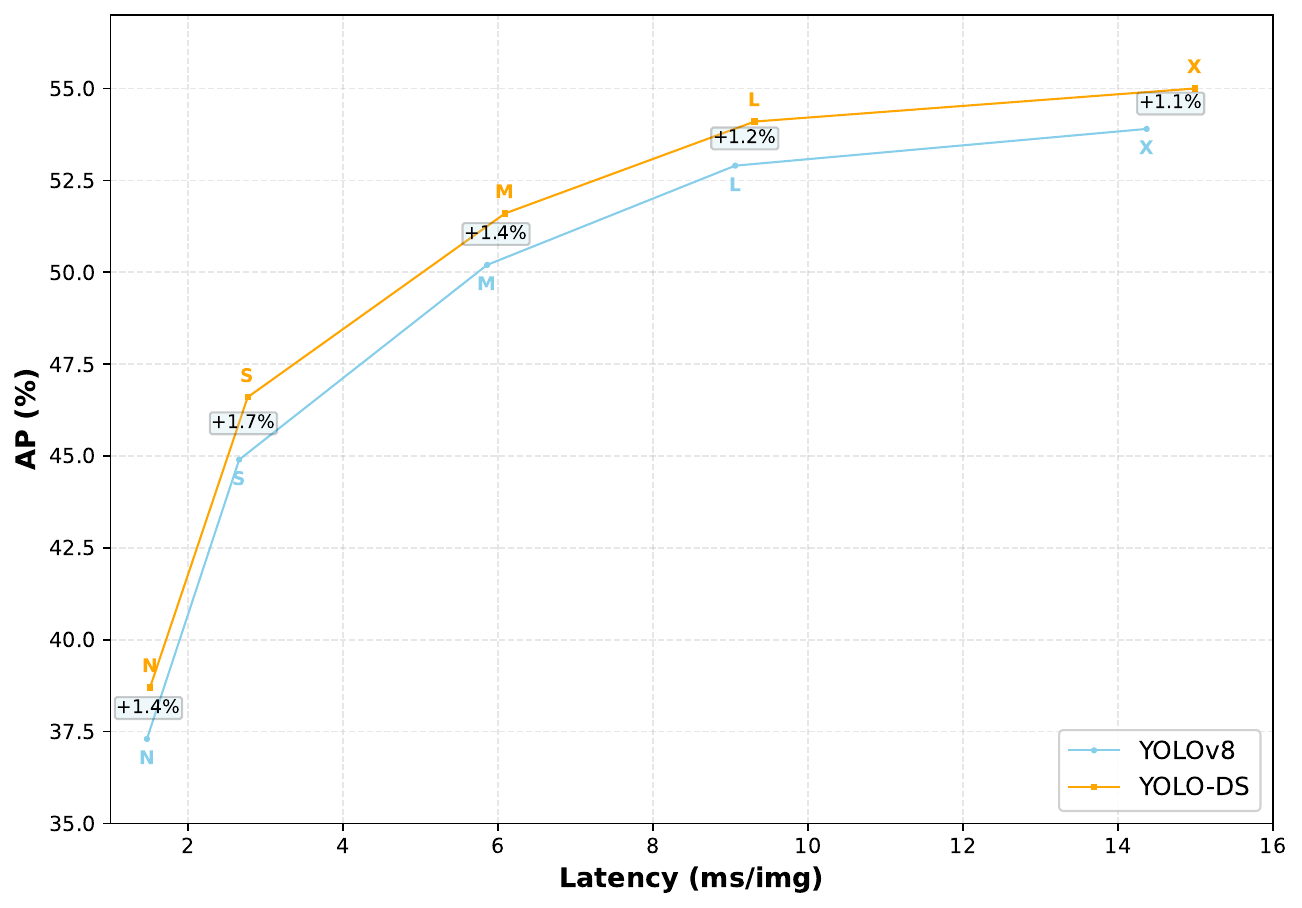}
        \centerline{(a)}
    \end{minipage}
    \begin{minipage}[b]{0.48\linewidth}
        \centering
        \includegraphics[scale=0.31]{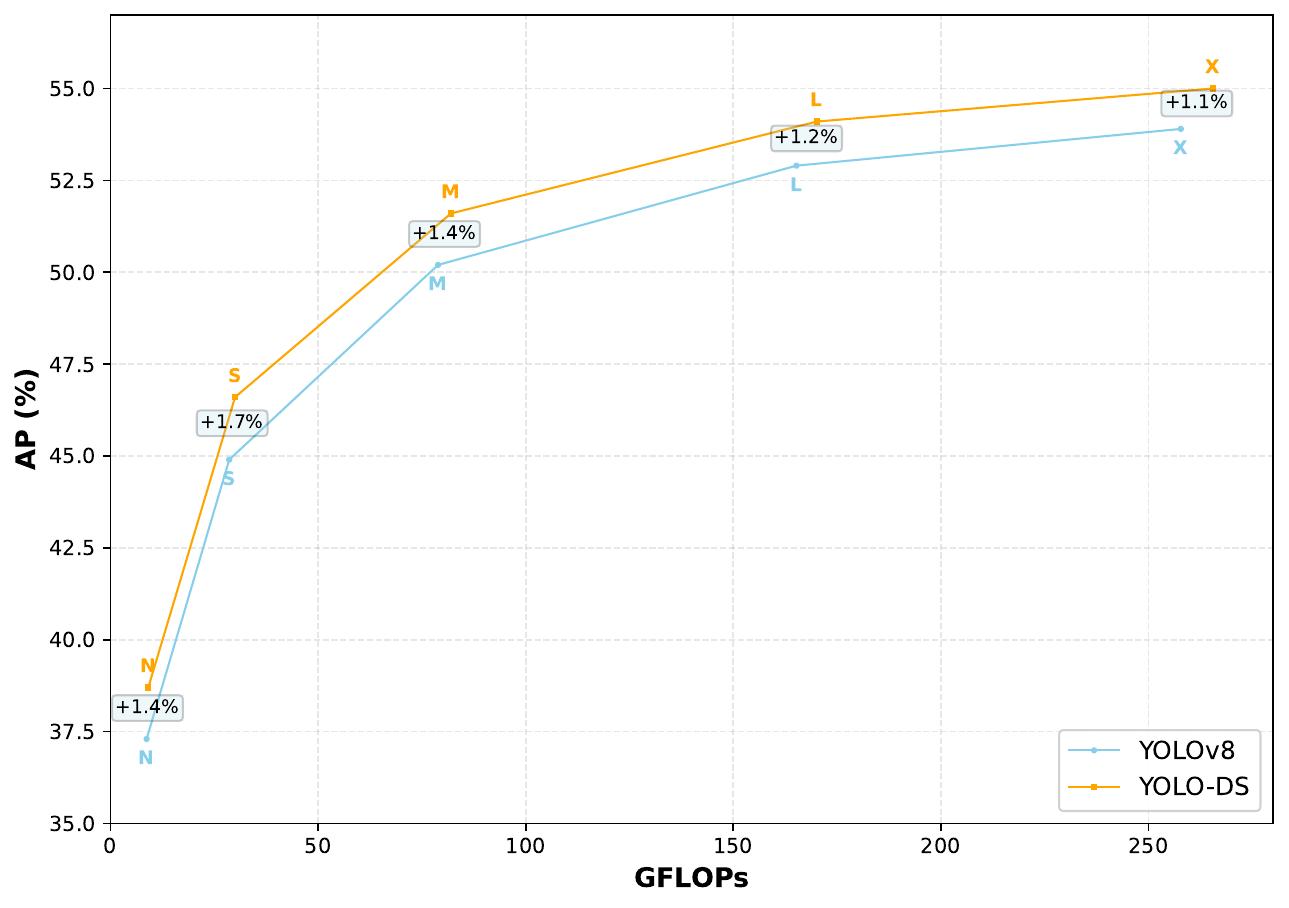}
        \centerline{(b)}
    \end{minipage}
    \caption{Comparison of the proposed YOLO-DS with YOLOv8.}
    \label{fig:data}
\end{figure}

\iffalse
%尽管SENet、CBAM以及视觉Transformer中的多头自注意力（MHSA）等现有注意力机制已取得成功，它们在建模目标特定响应分布方面仍存在根本性局限。SENet仅依赖全局均值统计进行通道重校准，这难以充分表征由不同目标尺度引发的异构激活模式。CBAM在SENet基础上进一步引入空间注意力机制，通过结合空间平均统计与最大统计量，但其全局空间聚合操作不可避免地混合了来自多个目标的激活响应，从而模糊了细粒度的通道语义。MHSA虽然通过多个注意力头建模全局依赖关系，但每个头内部的特征交互仍然存在隐式纠缠，导致其中小目标或稀疏目标的响应极易被淹没。此外，追求更细粒度的头部分解会导致计算成本的二次增长，使其难以适用于实时检测场景。

Despite their success, existing attention mechanisms such as SENet\cite{senet}, CBAM\cite{cbam}, and multi-head self-attention (MHSA) in ViT\cite{vit}  exhibit fundamental limitations in modeling object-specific response distributions. SENet relies exclusively on global mean statistics for channel recalibration, which inadequately represents heterogeneous activation patterns induced by diverse object scales. Building upon SENet, CBAM further introduces spatial attention by exploiting both spatial average and max statistics, yet its global spatial aggregation inevitably mixes activation responses from multiple objects, thereby obscuring fine-grained channel semantics. Although MHSA models global dependencies via multiple attention heads, feature interactions within each head remain implicitly entangled, where responses from small or sparse objects are prone to being overwhelmed. Moreover, pursuing finer-grained head-wise decomposition leads to a quadratic increase in computational cost, making it less suitable for real-time detection.
\fi

While attention mechanisms like SENet\cite{senet} and CBAM\cite{cbam} attempt to recalibrate features, they exhibit critical shortcomings for this task. SENet relies exclusively on global mean statistics for channel recalibration, which inadequately represents heterogeneous activation patterns induced by diverse object scales. Building upon SENet, CBAM further introduces spatial attention by exploiting both spatial average and max statistics, yet its global spatial aggregation inevitably mixes activation responses from multiple objects, thereby obscuring fine-grained channel semantics. Although multi-head self-attention (MHSA) in ViT\cite{vit} models global dependencies via multiple attention heads, feature interactions within each head remain implicitly entangled, where responses from small or sparse objects are prone to being overwhelmed. Moreover, pursuing finer-grained head-wise decomposition leads to a quadratic increase in computational cost, making it less suitable for real-time detection.

To bridge this gap, we propose YOLO-DS. Our core contribution is the \textbf{Dual-Statistic Synergy Operator (DSO)}, which explicitly models the channel-wise mean ($\mu$, representing overall strength) and the peak-to-mean difference ($d$, representing sparsity/saliency) in a synergistic 2D decision space\cite{ds}. This allows for the natural decoupling of feature responses into categories such as small objects, large objects, mixed-scale objects and background. Leveraging the DSO, we further introduce:

1. The \textbf{Dual-Statistic Synergy Gating (DSG)} Module: A channel gating mechanism that uses DSO outputs to generate adaptive weights, enhancing discriminative features and suppressing noise.

2. The \textbf{Multi-Path Segmented Gating (MSG)} Module: A depth-wise gating mechanism that dynamically weights features from different network depths within the C2F structure, optimizing representations for objects of varying scales.

Integrated into the YOLO architecture, YOLO-DS delivers consistent and scalable improvements. On MS-COCO\cite{mscoco}, it outperforms YOLOv8\cite{yolo8} across all model scales (N, S, M, L, X) by 1.1–1.7\% AP\cite{mscoco} (Fig.\ref{fig:data}) with negligible latency overhead.
﻿
﻿

%1. 基于解析门控的细粒度全通道特征解耦
%我们提出一种基于非参数化、可解释的通道门控算子的细粒度全通道门控模块。该算子将通道响应统计量显式映射到一个高维决策空间。通过细粒度特征解耦，所提出的算子能有效区分异质目标，并在特征融合前生成目标感知的门控信号。该机制自适应地增强与目标相关的激活，同时抑制冗余响应，仅需极少的额外参数量和浮点运算量即可实现显著的检测性能提升。

%2. 具备稳健权重调节的多路径分段细化门控
%我们进一步引入一种多路径分段细化门控模块，该模块在特征聚合前，对不同深度和融合路径的特征进行分组权重调制。通过融入多维温度调节机制和高斯分布噪声扰动，该模块提升了门控权重的鲁棒性与稳定性。尽管具备精细的控制能力，该模块仅产生微小的计算开销，从而在有限复杂度条件下实现了持续性的精度改进。

%We further introduce a multi-path segmented refinement gating module, which performs grouped weight modulation over features from different depths and fusion paths before aggregation. By incorporating a multi-dimensional temperature control mechanism and Gaussian-distributed noise perturbation, the proposed module improves the robustness and stability of gating weights. Despite its refined control capability, the module incurs only marginal computational overhead, leading to consistent accuracy improvements under constrained complexity.

\section{Methodology}

%\subsection{Fine-Grained Feature Decoupling Operator (AO)}
\subsection{Dual-Statistic Synergy Operator (DSO)}

%我们首先系统分析了 SENet、CBAM 以及多头自注意力机制在建模异构目标时所存在的局限性。进一步地，通过对不同类型目标在特征图中的通道响应分布进行统计分析，我们发现 仅依赖通道均值或通道最大值难以有效区分不同目标类型。基于此，本文采用 通道均值与峰均差（peak-to-mean difference） 作为两个互补的统计维度，对不同目标的通道响应特征进行刻画。其中，通道均值用于衡量整体激活强度，而峰均差则刻画通道响应的离散程度，用以反映局部强激活与全局弱响应之间的差异。基于这两个统计量的联合建模，我们将目标表征划分为小目标、大目标、大小目标共存以及无目标四种典型模式。设𝑎表示通道均值，b 表示通道最大值，其具体计算过程如下：

We first conduct a systematic analysis of the limitations of SENet, CBAM, and multi-head self-attention mechanisms\cite{vit} in modeling heterogeneous objects. Further statistical analysis of channel response distributions reveals that neither the channel-wise mean nor the maximum activation alone is sufficient to distinguish different object types. To this end, we adopt the channel-wise mean ($\mu$) and the peak-to-mean difference ($d$) as two complementary statistical descriptors for characterizing channel responses. Specifically, the channel-wise mean reflects the overall activation strength, while the peak-to-mean difference captures the degree of response sparsity, highlighting the discrepancy between localized strong activations and globally weak responses. By jointly modeling these two statistics, object representations can be categorized into four canonical patterns: small objects, large objects, mixed-scale objects, and background (no-object) responses. Assuming the input feature map is $x\in\mathbb{R}^{B \times C \times H \times W}$. The detailed computation is given as follows:
\begin{equation}
\mu_{b,c} = \frac{1}{HW} \sum_{h=1}^{H} \sum_{w=1}^{W} x_{b,c,h,w}, \;\;\;\; \mu\in\mathbb{R}^{B \times C \times 1 \times 1}
\end{equation}
\begin{equation}
m_{b,c} = \max_{h,w} x_{b,c,h,w}, \;\;\;\; m\in\mathbb{R}^{B \times C \times 1 \times 1}
\end{equation}
\begin{equation}
d_{b,c} = m_{b,c}- \mu_{b,c}, \;\;\;\; d\in\mathbb{R}^{B \times C \times 1 \times 1}
\end{equation}
% 其中d是峰均差。我们希望根据峰均差和均值两个统计量的相互变化影响，设计了一种能否反应二维变化的算子，具体如下：
where $\mu_{b,c}$ and $m_{b,c}$ denote the channel-wise mean and maximum response, respectively, and $d_{b,c}$ represents the peak-to-mean difference. Rather than treating the two statistics independently, we aim to explicitly model the coupled effects of the channel-wise mean and the peak-to-mean difference within a two-dimensional decision space:
\begin{equation}
s_{b,c} = (\mu_{b,c},d_{b,c})\in\mathbb{R}^2
\end{equation}
where $\mu_{b,c}$ characterizes the overall activation strength, and $d_{b,c}$ reflects the degree of response sparsity and localized saliency. To capture the coupled relationship between overall activation strength ($\mu$) and response sparsity ($d$), we design the DSO as a synergistic function:
\begin{equation}
\Phi(\mu, d) = (d + 1)(\mu + 1) - 1
\end{equation}
which can be equivalently expanded as
\begin{equation}
\Phi(\mu, d) = \mu d + \mu +d
\label{eq:ao}
\end{equation}
\begin{equation}
y_{b,c} = \Phi(\mu_{b,c}, d_{b,c}), \;\;\;\; y\in\mathbb{R}^{B \times C \times 1 \times 1}
\end{equation}
%其中，𝑦表示由我们的分析算子产生的通道决策响应。该算子可分解为通道均值 μ、峰均差 Δ 及其相互作用的贡献。
where $y_{b,c}$ denote the channel-wise decision response produced by our DSO. 
%The operator decomposes into contributions from the channel mean $\mu$, the peak-to-mean difference $d$, and their interaction. 
%为了对DSO进一步分析，我们对等式6进行偏导数计算，具体计算过程如下：
To gain deeper insight into the DSO, we compute the partial derivatives of Eq.\ref{eq:ao} as follows:
\begin{equation}
\frac{\partial\Phi}{\partial\mu} = d+1 \textgreater 0, \frac{\partial\Phi}{\partial d} = \mu+1 \textgreater  0
\end{equation}
Therefore, under the constraint of non-negative feature activations the DSO $\Phi(\mu, d)$ is strictly monotonically increasing with respect to both $\mu$ and $d$. This indicates that stronger global activation (an increase in $\mu$) or greater local sparsity (an increase in $d$) will both lead to an increase in the value of $\Phi$. Further compute the second-order partial derivatives:
\begin{equation}
\frac{\partial}{\partial d}(\frac{\partial\Phi}{\partial\mu}) = \frac{\partial}{\partial d}(d+1)=1 \textgreater 0,
\frac{\partial}{\partial\mu}(\frac{\partial\Phi}{\partial d}) = \frac{\partial}{\partial\mu}(\mu+1)=1 \textgreater  0
\label{eq:ao2}
\end{equation}
%Eq.9 shows a larger $d$ corresponds to a greater marginal contribution of the $\mu$ to $\Phi$. Eq.10 shows a larger $\mu$ corresponds to a greater marginal contribution of the $d$ to $\Phi$. Our analysis reveals a synergistic effect between $\mu$ and $d$.
Based on the analysis of Eq.\ref{eq:ao2}, it can be observed that a larger value of $d$ enhances the marginal contribution of $\mu$ to $\Phi$, while a larger $\mu$ similarly increases the marginal contribution of $d$ to $\Phi$. This indicates a synergistic relationship between $\mu$ and $d$, where each variable reinforces the contribution of the other to the overall response $\Phi$.

\begin{figure}[H]
\centering
    \begin{minipage}[b]{0.48\linewidth}
        \centering
        \includegraphics[scale=0.43]{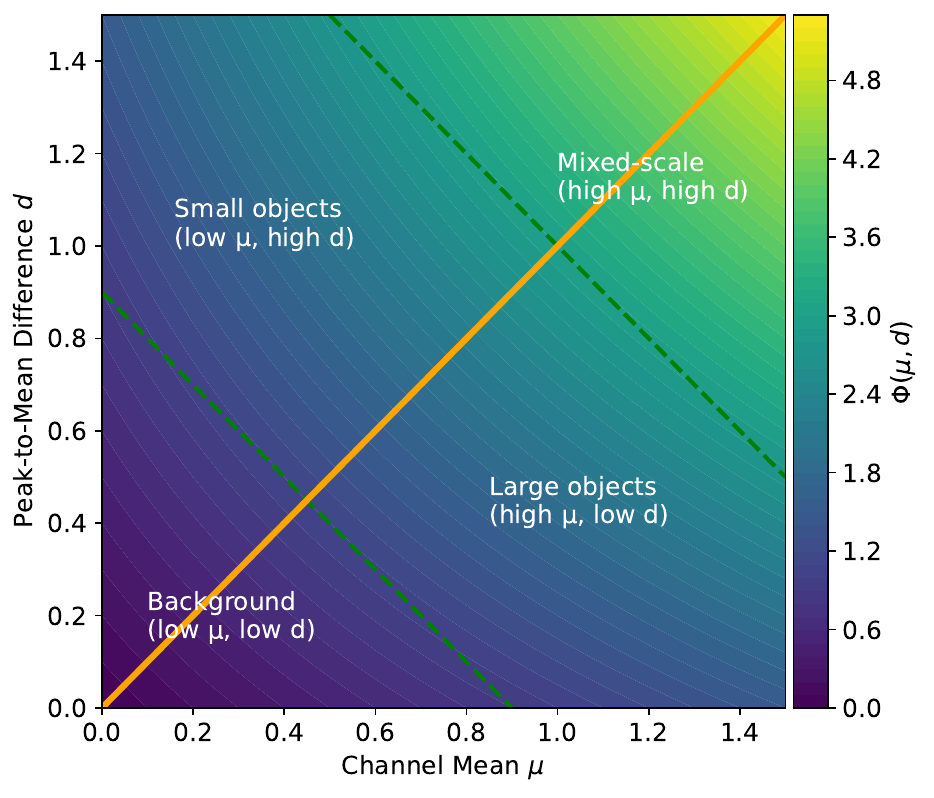}
        \centerline{(a)}
    \end{minipage}
    \begin{minipage}[b]{0.48\linewidth}
        \centering
        \includegraphics[scale=0.43]{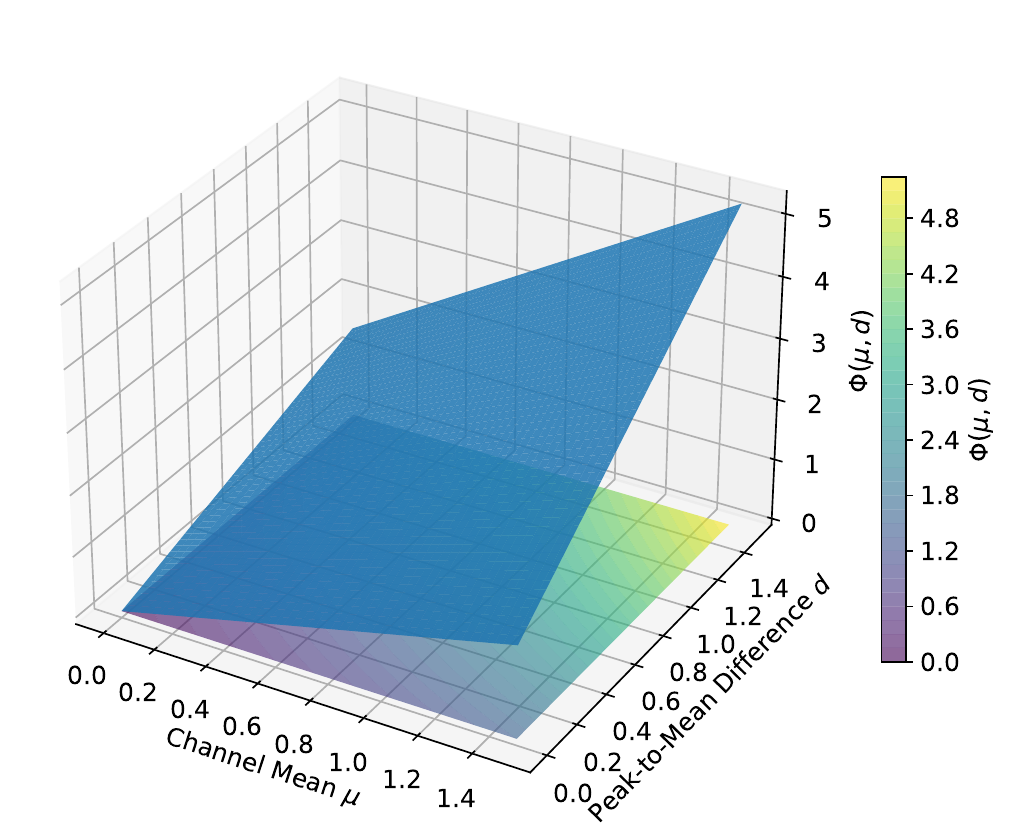}
        \centerline{(b)}
    \end{minipage}
    \caption{(a) Two-dimensional contour map of the DSO response as a function of $\mu$ and $d$. (b) Three-dimensional rendering of $\Phi(\mu, d)$, explicitly showing the nonlinear surface shaped by the interplay of activation strength and sparsity.}
    \label{fig:ao}
\end{figure}

%为了更深入地理解所提分析算子（DSO）的行为特性，我们将其响应以二维与三维形式进行可视化呈现，分别对应图1(a)与图1(b)。具体而言，图1(a)展示了以通道均值μ和峰均差d为基轴展开的二维决策空间，其中的等高线图谱描绘了DSO触发的非线性响应。图1(b)进一步呈现了对应的三维决策曲面，清晰揭示了全局激活强度与局部响应稀疏性之间的耦合交互如何调控通道级的决策响应。
To gain deeper insight into the behavior of the proposed DSO, we visualize its response in both two-dimensional and three-dimensional forms, as shown in Fig.\ref{fig:ao} (a) and Fig.\ref{fig:ao} (b), respectively. 
%图1(a)中的橙色实线 d=μ（对应 m=2μ）为响应区域提供了一个有效的划分边界。该边界线随 d 和 μ 向正无穷方向延伸，其对应的激活强度严格单调递增。因此，位于该橙色实线上方的区域（即 d>μ，m>2μ）可定义为稀疏显著（或局部聚焦）区域，其中最大激活值高于平均值。这表明特征图中仅少数空间位置（例如物体的关键点或小物体的中心）被强烈激活，而大部分区域响应微弱，激活分布表现出高度的不均匀性与稀疏性。该区域通常对应小目标、边缘、纹理细节等显著的局部特征。相反，位于该橙色实线下方的区域（即 d<μ，m<2μ）可定义为均匀广泛（或全局弥散）区域，其中最大激活值与平均值的差异较小。这意味着特征图的激活在空间上分布相对均匀、广泛，不存在尖锐的峰值，所有位置对响应的贡献较为平均。该区域一般对应大目标、背景或对比度较低的大范围分布特征。
The orange solid line in Fig.\ref{fig:ao} (a), given by $d=\mu$ (corresponding to $m=2\mu$), provides an effective boundary that partitions the response region. As this orange boundary extends toward positive infinity in both $d$ and $\mu$, the corresponding activation strength increases strictly monotonically. Therefore, the region above this orange line (i.e., $d \textgreater \mu,m \textgreater 2\mu$) can be defined as the sparse-salient (or locally focused) region, where the maximum activation exceeds the mean. This indicates that only a small number of spatial locations in the feature map—such as keypoints of objects or centers of small objects—are strongly activated, while most areas exhibit weak responses. The activation distribution is highly non-uniform and sparse. This region typically corresponds to salient local features such as small objects, edges, and fine texture details. Conversely, the region below the orange line (i.e.,  $d \textless \mu,m \textless 2\mu$) can be defined as the uniform-broad (or globally diffuse) region, where the difference between the maximum activation and the mean is relatively small. This implies that activations are distributed more evenly and extensively across spatial locations, without sharp peaks, and all positions contribute relatively uniformly to the response. This region generally corresponds to large objects, background, or broadly distributed features with low contrast.

%为更详细的区分不同的目标，我们引入了两条非确定边界（图1中绿色虚线），把所有响应\Phi划分为4个模糊区域，用于区分小目标、大目标、混合目标及背景，针对这四个区域具体界定如下：1、小目标： 高 d， 相对低 μ，因此 d  > μ，即图1中的左上角区域。2、大目标： 高 μ， 相对低 d，因此 d < μ，即图1中的右下角。3、混合尺度目标： 同时包含局部强响应和广泛弱响应，其统计量 μ 和 d 都处于较高值，即沿 d ≈ μ方向且\Phi更强响应处波动，即图1中的右上角。4、背景： μ 和 d 都较低，即沿 d ≈ μ方向且\Phi更弱响应处波动，即图1中的左角。
To clearly distinguish between different objects, we introduce two non-deterministic boundaries (green dashed lines in Fig.\ref{fig:ao} (a)) to partition all responses $\Phi$ into four fuzzy regions, which are used to differentiate small objects, large objects, mixed-scale objects, and background. These four regions are specifically defined as follows:
\begin{itemize}
\item Small Objects: Characterized by a high value of $d$ and a relatively low value of $\mu$, satisfying $d \textgreater \mu,m \textgreater 2\mu$. This corresponds to the upper-left region in Fig.\ref{fig:ao} (a).

\item Large Objects: Characterized by a high value of $\mu$ and a relatively low value of $d$, satisfying $d \textless \mu,m \textless 2\mu$. This corresponds to the lower-right region in  Fig.\ref{fig:ao} (a).

\item Mixed-Scale Objects: Exhibit both locally strong responses and broadly weak responses, with both statistical measures $\mu$ and $d$ taking relatively high values. These objects fluctuate approximately along the direction where $d \approx \mu$, in regions where $\Phi$ shows stronger responses, corresponding to the upper-right area in  Fig.\ref{fig:ao} (a).

\item Background: Both $\mu$ and $d$ are low, fluctuating approximately along the direction where $d \approx \mu$, in regions where $\Phi$ shows weaker responses, corresponding to the lower-left area in  Fig.\ref{fig:ao} (a).
\end{itemize}
%我们引入的两条非确定边界所确定的4个模糊区域并不是固定的，仅为可视化分析提供视觉案例，实际需要通过学习输入样本的特征分布进一步确认，而且这4个分类目标中除了背景分类都会有不同程度的噪声干扰，比如小目标的分类中可能会有稀疏高响应噪点干扰，大目标分类中可能会出现大面积背景噪声干扰，因此仍然需要通过学习不同样本特征对其进行过滤。因此，总的来说我们提出的算子DSO是在门控网络学习特征前对不同目标的特征解耦以及对目标分类的预处理，并通过这种方式使门控网络更好的学习不同目标特征，进而进行更有效的分类，以自适应的方式动态保留高激活特征，剔除冗余特征，提升目标检测网络检测精度。
Except for one decision boundary ($d=\mu, m=2\mu$, orange solid line in Fig.\ref{fig:ao} (a)) which is fixed, the four fuzzy regions determined by the two non-deterministic boundaries (green dashed lines in Fig.\ref{fig:ao} (a)) we introduce are not rigid. They primarily serve as a visual schematic for analysis. In practice, their precise delineation must be further refined by learning the feature distribution from input samples. Furthermore, among these four categories, all except the background are susceptible to varying degrees of noise interference. For example, the small-object category may be perturbed by sparse, high-response noise points, while the large-object category could be affected by extensive background noise. Consequently, filtering via learning from diverse sample features remains essential.

In summary, the proposed DSO acts as a feature decoupling mechanism for different objects and a preprocessing step for object classification before the gating network learns the features. By doing so, it enables the gating network to better learn the distinctive features of different objects, thereby facilitating more effective classification. In an adaptive manner, it dynamically retains highly activated features while discarding redundant ones, ultimately enhancing the detection accuracy (AP) of the object detection network.

\subsection{Dual-Statistic Synergy Gating (DSG)}

%尽管双统计协同算子（DSO）能够解耦不同对象的特征，并作为对象分类的预处理步骤，但由此得到的特征仍然容易受到两种类型的干扰：跨类别特征混合和噪声污染。为了解决这些问题，我们在DSO之后引入了一个$1 \times 1$的卷积层。这个可训练的卷积层通过增强判别信号并抑制预分类组内的类间干扰和噪声，进一步净化特征表示。随后，一个sigmoid函数激活了细化后的特征图。特征随后由得到的0-1权重进行门控，其中冗余或噪声激活被抑制，有信息的激活被选择以实现注意力门控机制。我们将这个模块称为双统计协同门控（DSG）网络（图\ref{fig:dsg}中的蓝色区域）。其主要计算过程如下：

\begin{figure}[H]
  \centering
  \includegraphics[scale=0.48]{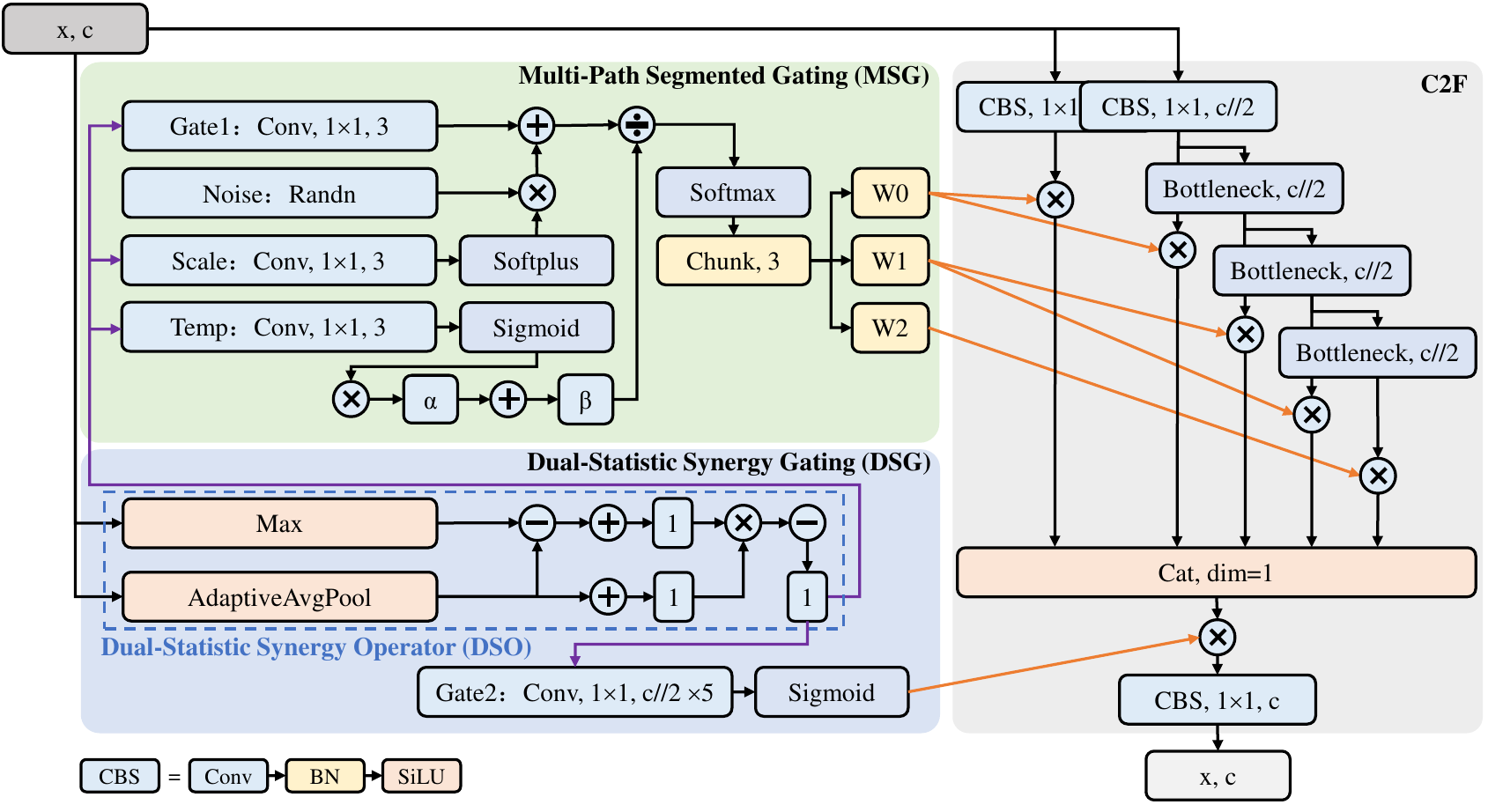}
  \caption{Structural diagram of the C2F architecture incorporating the DSG and MSG modules.}
  \label{fig:dsg}
\end{figure}

Although the DSO decouples features of different objects and serves as a preprocessing step for object classification, the derived features remain susceptible to two types of interference: cross-category feature mixing and contamination by noise. To address these issues, we introduce a $1 \times 1$ convolutional layer after DSO. This trainable convolution further purifies the feature representations by enhancing discriminative signals and suppressing both inter-class interference and noise within the pre-classified groups. Subsequently, a sigmoid function activates the refined feature maps. Features are then gated by the resulting 0–1 weights, where redundant or noisy activations are suppressed, and informative ones are selected to implement an attentional gating mechanism. We term this module the Dual-Statistic Synergy Gating (DSG) network(the blue region in Fig.\ref{fig:dsg}). Its main computational procedure is as follows:
\begin{equation}
\begin{gathered}
z_{DSG} = \mathbf{W}_{DSG} * y + \mathbf{b}_{DSG}, \;\;\;\; z_{DSG}\in\mathbb{R}^{B \times C' \times 1 \times 1}, \\ \mathbf{W}_{DSG}\in\mathbb{R}^{C' \times C \times 1 \times 1}, \mathbf{b}_{DSG}\in\mathbb{R}^{C'}, C'=\lfloor \frac{C}{2} \rfloor \times (2+n) 
\end{gathered}
\end{equation}
\begin{equation}
w_{DSG} = \sigma(z_{DSG}) = \frac{1}{1+e^{-z_{DSG}}}, \;\;\;\; w_{DSG}\in\mathbb{R}^{B \times C' \times 1 \times 1}
\end{equation}
\begin{equation}
x_{out} = w_{DSG} \odot x_{cat}, \;\;\;\; x_{out}\in\mathbb{R}^{B \times C' \times H \times W},x_{cat}\in\mathbb{R}^{B \times C' \times H \times W}
\end{equation}
%其中n为瓶颈结构的数量，即网络深度，W为输入通道C输出通道为C'的1×1卷积，b为通道数C’的偏置项，sigma为sigmoid函数，xcat为C2F结构中特征concat处，xout为最终输出结果。
where  $\mathbf{W}_{DSG}$ represents the $1\times1$ convolutional kernel transforming $C$ input channels to $C'$ output channels, $\mathbf{b}_{DSG}$ is the bias vector of dimension $C'$, $n$ denotes the number of bottleneck blocks\cite{resnet}, $\sigma$ is the sigmoid activation function, $x_{\text{cat}}$ corresponds to the feature concatenation point in the C2F structure\cite{yolo8}, and $x_{\text{out}}$ is the final output.

%在C2F结构的基础上增加DSG模块和MSG模块的结构图

%总体而言，该门控模块是对DSO的补充。它通过不同输入样本间双重统计量（$d$和$\mu$）的协同作用，有效地解耦了异构对象特征。此外，它还以可学习的方式进一步过滤跨类别特征和干扰噪声，自适应地形成更具区分性的异构对象表示，进而生成有效的门控权重。这些权重在C2F结构中信息密集的拼接点高效地调节多通道特征，在保留信息特征的同时抑制冗余特征。该机制成功解耦了小目标、大目标、混合尺度和背景的特征，在计算开销极小的情况下显著提高了平均精度（AP）。

Overall, this gating module serves as a complementary addition to the DSO. It effectively decouples heterogeneous object features through the synergistic interaction of the dual statistics ($d$ and $\mu$) across different input samples. Furthermore, in a learnable manner, it further filters cross-category features and interference noise, adaptively forming more discriminative heterogeneous object representations, which in turn produce effective gating weights. These weights efficiently regulate the multi-channel features at the information-intensive concatenation point in the C2F structure, preserving informative features while suppressing redundant ones. This mechanism successfully decouples the features of small objects, large objects, mixed-scale, and background, leading to significant gains in AP with minimal computational overhead.

\subsection{Multi-Path Segmented Gating (MSG)}
% 在C2F模块中，多个瓶颈块按顺序堆叠，并将它们的中间输出进行拼接。尽管这些特征共享相同的空间分辨率，但由于计算路径长度的差异，对应的有效网络深度是不同的。因此，不同拼接路径的特征在语义抽象和感受野方面表现出不同的表示偏好。研究表明，早期瓶颈输出保留了细粒度的空间细节和局部结构，这有利于小目标特征提取；而深层瓶颈输出则编码了更抽象的语义和上下文信息，更适用于大目标特征提取和背景抑制。因此，对所有瓶颈输出采取同等处理方式导致特征利用效果欠佳。

In the C2F module, multiple bottleneck blocks are stacked sequentially, and their intermediate outputs are concatenated. Although these features share the same spatial resolution, their corresponding effective network depths differ due to varying computational path lengths. Consequently, features from different concatenation paths exhibit distinct representational biases in terms of semantic abstraction and receptive field. It has been shown that early bottleneck outputs preserve fine-grained spatial details and local structures, which facilitates feature extraction for small objects, whereas deeper bottleneck outputs encode more abstract semantic and contextual information, making them more suitable for large-object feature extraction and background suppression\cite{convana,convana2,convana3}. Therefore, treating all bottleneck outputs equally leads to suboptimal feature utilization.

%我们在DSO和DSG模块中，对异构目标进行有效区分，但是，对于不同网络深度对异构目标特征提取的重要程度仍然需要进一步区分，以达到进一步提升AP的目标。作为补充，我们对C2F结构中不同的路径进行了分组， 并通过softmax来平衡不同分组权重，以自适应的方式实现不同尺度目标对于网络层深特征的竞争性需求。为了进一步提升分组权重的鲁棒性，我们还引入了可自适应缩放的符合高斯分布的随机噪声，以及可以更灵活控制权重之间剧烈程度的自适应多维控温模块，并命名为Multi-Path Segmented Gating (MSG)，如图2绿色部分所示。该模块具体计算过程如下所示：
Although the proposed DSO and DSG modules effectively discriminate heterogeneous objects, the relative importance of different network depths in extracting discriminative features for heterogeneous objects has not yet been sufficiently explored, which limits further improvements in detection performance. To address this issue, we further group different computational paths within the C2F structure according to their effective network depths, and employ a softmax-based weighting strategy to balance the contributions of different groups. This design enables an adaptive competition among depth-level features, allowing objects of different scales to dynamically select suitable representations from shallow to deep network layers. Moreover, to enhance the robustness of the grouping weights, we introduce an adaptively scaled Gaussian-distributed random noise, together with an adaptive multi-dimensional temperature control module that flexibly regulates the sharpness of the weight distribution. The resulting module is termed Multi-Path Segmented Gating (MSG), as illustrated by the green region in Fig.\ref{fig:dsg}. The detailed computational procedure of MSG is described as follows:
\begin{equation}
\begin{gathered}
z_{MSG} = \mathbf{W}_{MSG} * y + \mathbf{b}_{MSG}, \;\;\;\; z_{MSG}\in\mathbb{R}^{B \times 3 \times 1 \times 1}, \\ \mathbf{W}_{MSG}\in\mathbb{R}^{3 \times C \times 1 \times 1}, \mathbf{b}_{MSG}\in\mathbb{R}^{3} 
\end{gathered}
\end{equation}
\begin{equation}
\begin{gathered}
z_{scale} = \mathbf{W}_{scale} * y + \mathbf{b}_{scale}, \;\;\;\; z_{scale}\in\mathbb{R}^{B \times 3 \times 1 \times 1}, \\ \mathbf{W}_{scale}\in\mathbb{R}^{3 \times C \times 1 \times 1}, \mathbf{b}_{scale}\in\mathbb{R}^{3} 
\end{gathered}
\end{equation}
\begin{equation}
z_{noise} = ln(1+e^{z_{scale}}) \odot \epsilon,  \;\;\;\;  \epsilon \sim \mathcal{N}(0,I),\epsilon\in\mathbb{R}^{B \times 3 \times 1 \times 1}
\end{equation}
\begin{equation}
\begin{gathered}
\mathcal{T} =\alpha \cdot \sigma(\mathbf{W}_{t} * y + \mathbf{b}_{t}) +\beta = \alpha \cdot \frac{1}{1+e^{-(\mathbf{W}_{t} * y + \mathbf{b}_{t})}}+\beta,\\ \mathcal{T}\in\mathbb{R}^{B \times 3 \times 1 \times 1},  \mathbf{W}_{t}\in\mathbb{R}^{3 \times C \times 1 \times 1}, \mathbf{b}_{t}\in\mathbb{R}^{3} 
\end{gathered}
\end{equation}
\begin{equation}
\begin{gathered}
w_{MSG} = \mathbf{Softmax}(\frac{z_{MSG} + z_{scale}}{\mathcal{T}}), \;\;\;\; w_{MSG}\in\mathbb{R}^{B \times 3 \times 1 \times 1}
\end{gathered}
\end{equation}
%其中W_MSG表示门控部分的1*1卷积权重，b_MSG是对应的偏置项，在这里我们的权重分组是3，后面实验部分将作具体说明，z_MSG表示门控。W_scale表示噪声自适应缩放部分的卷积权重，b_scale是对应的偏置项，z_scale表示自适应噪声缩放，epsilon表示符合高斯分布的随机噪声，z_noise表示自适应随机噪声。W_t表示自适应多维温控模块的1×1卷积，b_t是对应的偏置项，alpha和beta是自适应多维温控模块的缩放参数，实验表明alpha=1.9，beta=0.1时效果最好，T表示自适应多维温控项。w_MSG是最终的分组权重。
where $\mathbf{W}_{MSG}$ and $\mathbf{b}_{MSG}$ denote the weight and bias of the $1 \times 1$ convolution in the gating component, respectively. The weights are divided into three groups, the rationale for which will be detailed in the experimental section. The resulting gating signal is denoted as $z_{MSG}$. For the adaptive noise scaling pathway, $\mathbf{W}_{scale}$ and $\mathbf{b}_{scale}$ represent the convolutional weight and bias, yielding the scaling factor $z_{scale}$. A Gaussian-distributed random noise $\epsilon$ is then modulated by $z_{scale}$ to produce the adaptive random noise $z_{noise}$. In the adaptive multi-dimensional temperature control module, $\mathbf{W}_t$ and $\mathbf{b}_t$ correspond to the weight and bias of another $1 \times 1$ convolution. The module employs two fixed scaling parameters, $\alpha$ and $\beta$, which are set to 1.9 and 0.1 respectively, as this configuration was found to yield the best performance in our experiments. The output of this module is denoted as the adaptive multi-dimensional temperature term $\mathcal{T}$. Finally, the grouped weighting coefficients $\mathbf{W}_{MSG}$ are obtained by normalizing the combined gating and temperature-adjusted signals through a softmax operation.

%总的来说，msg模块在c2f中不同路径特征concat前实现了对不同深度的特征进行分组，并通过自适应的方式调节不同深度网络权重，通过这种方式能进一步从网络深度的层面有效优化大目标和小目标的特征权重进而增加很小的计算量同时提升ap，这对dso和dsg模块对异构目标区分来说是从网络深度维度的有效补充。
Overall, the MSG module groups features of varying network depths within the C2F structure before their concatenation and adaptively modulates the weights assigned to different depth levels. This design enables a further, effective optimization of feature contributions for both large and small objects from the perspective of network depth. Consequently, it enhances AP with only minimal computational overhead. This approach provides a complementary enhancement to the DSO and DSG modules in distinguishing heterogeneous objects by introducing an adaptive gating mechanism along the depth dimension.

\section{Experiments}
%为评估模型性能，所有实验均在配备8张NVIDIA GeForce RTX 4090d GPU的工作站上进行。训练使用MS COCO 2017数据集，该数据集包含约118,287张训练图像和5,000张验证图像，涵盖80个目标类别，是衡量目标检测算法性能的通用基准。对比实验的基线模型（Baseline）选用广泛应用的YOLOv8系列（包括n, s, m, l, x变体），改进模型为YOLO-DS系列（包括n, s, m, l, x变体）。为公平比较，所有模型均在640×640像素输入分辨率、相同数据增强策略下，使用官方代码库和推荐超参数从头训练（train from scratch）至收敛。评估指标采用目标检测领域通用的平均精度（mean Average Precision, mAP），包括mAP@50-95（IoU阈值从0.5到0.95的平均值）和mAP@50，同时记录模型参数量（Parameters）、计算量（FLOPs）以及在T4 GPU上使用TensorRT加速的推理速度（延迟）。

To evaluate model performance, all experiments were conducted on a workstation equipped with eight NVIDIA GeForce RTX 4090 GPUs. Training was performed using the MS COCO 2017 dataset\cite{mscoco}, which comprises approximately 118,287 training images and 5,000 validation images across 80 object categories, serving as a common benchmark for object detection algorithms. For comparative experiments, the baseline models were selected from the widely adopted YOLOv8 series (including the n, s, m, l, and x variants), while the improved models were from the YOLO-DS series (also including the n, s, m, l, and x variants). To ensure a fair comparison, all models were trained from scratch using the official codebases and recommended hyperparameters until convergence, with a consistent input resolution of 640×640 pixels and the same data augmentation strategies. Evaluation metrics followed the common practice in object detection, including mean Average Precision (AP)—specifically AP@50-95 (averaged over IoU thresholds from 0.5 to 0.95) —along with the number of parameters, FLOPs, and inference speed (latency) measured on a T4 GPU with TensorRT acceleration.

\subsection{Visualization Analysis}
%为了进一步探究所提出的YOLO-DS的有效性，我们采用Eigen-CAM来可视化不同检测器的注意力分布。Eigen-CAM直观地解释了模型如何在特征图上分配空间注意力，从而能够在不同的物体尺度和密度条件下，对特征利用和目标感知能力进行定性比较。

To further investigate the effectiveness of the proposed YOLO-DS, we employ Eigen-CAM\cite{eigen} to visualize the attention distributions of different detectors. Eigen-CAM provides an intuitive interpretation of how models allocate spatial attention across feature maps, enabling a qualitative comparison of feature utilization and target perception capability under various object scale and density conditions.
﻿

\begin{figure}[H]
  \centering
  \includegraphics[scale=0.7]{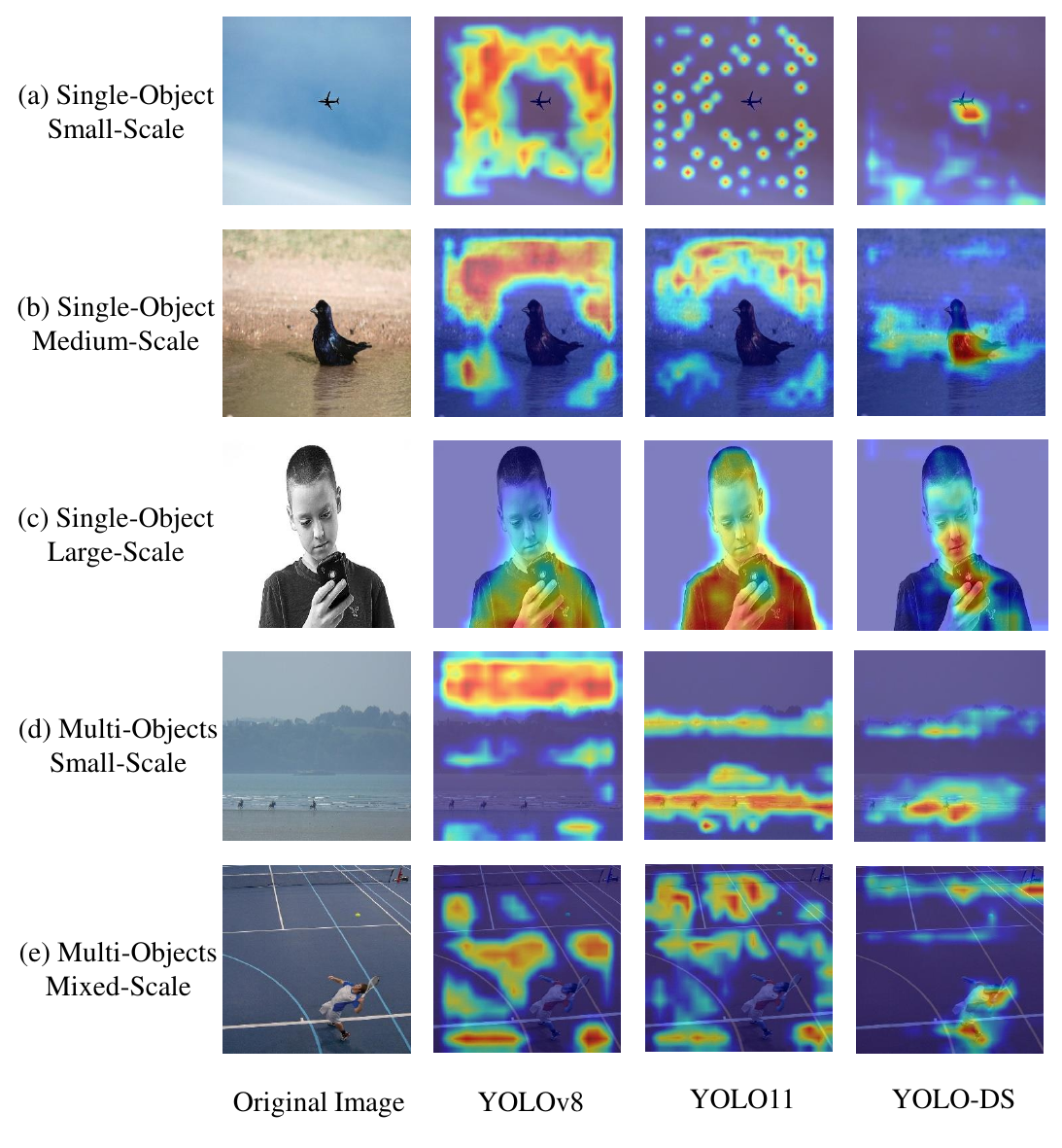}
  \caption{Eigen-CAM visualization of attention responses for different object-scale and object-density scenarios.}
  \label{fig:va}
\end{figure}

% 根据图3中所示，由于yolov8和yolo11并未对不同尺度目标进行有效区分，导致小目标检测场景出现了不同程度的注意力失焦现象（图（a）（b）（d）（e）），把更多的注意力激活响应分布在背景区域。尽管yolo11使用了多头自注意力机制，但分头的数量问题导致较小目标淹没在大目标或背景特征里，所以其对小目标的检测效果仍然较差。相比之下，YOLO-DS始终展现出更紧凑且更具区分度的激活图。对于单个小尺度对象，所提出的方法生成了尖锐且局部化的峰值，这表明其能够放大稀疏但显著的响应。对于大尺度对象，一个重要的观察结果是，YOLO-DS并非简单地在整个对象范围内最大化注意力（图（c））。与YOLOv8和YOLO11（其响应倾向于均匀覆盖大对象）相比，YOLO-DS产生的激活更具选择性，强调了区分性的局部结构和细粒度细节，如对象边界和特征语义区域。这一行为与我们的分析算子的设计相一致。当通道平均激活较高时，均匀强烈的响应通常表明特征冗余或信息量较少。通过综合考虑峰均差与平均值的差异，YOLO-DS抑制了过度的均匀激活，并将注意力重新分配到局部显著区域。因此，该模型聚焦于更为关键的细粒度细节，而非不加区分地放大整个对象区域。

%总体而言，YOLO-DS表现出一种尺度自适应的注意力行为：对于小目标，它会放大稀疏的峰值；而对于大目标，则会优先考虑具有区分性的细节，而非冗余区域。
As illustrated in Fig.\ref{fig:va}, YOLOv8 and YOLO11\cite{yolo11} fail to explicitly differentiate objects of varying scales, which leads to noticeable attention misalignment in small-object detection scenarios (Fig.\ref{fig:va}(a), (b), (d), and (e)). Specifically, a considerable portion of the activation responses is distributed over background regions rather than being concentrated on target objects. Although YOLO11 incorporates multi-head self-attention, the limited number of attention heads inevitably causes responses from small objects to be overwhelmed by large objects or background features, resulting in suboptimal performance on small-object detection. In contrast, YOLO-DS consistently produces more compact and discriminative activation maps across all scenarios. For single small-scale objects, the proposed method generates sharp and highly localized activation peaks, indicating its ability to amplify sparse yet salient responses.

For large-scale objects, an important observation is that YOLO-DS does not simply maximize attention over the entire object extent (Fig.\ref{fig:va}(c)). Compared with YOLOv8 and YOLO11, whose responses tend to uniformly cover large objects, YOLO-DS yields more selective activations that emphasize discriminative local structures and fine-grained details, such as object boundaries and semantically informative regions. This behavior is consistent with the design of our DSO. When the channel-wise mean activation is high, uniformly strong responses often indicate feature redundancy or limited discriminative power. By jointly considering the peak-to-mean difference and the mean activation, YOLO-DS suppresses excessive homogeneous responses and reallocates attention to locally salient regions. Consequently, the model focuses on critical fine-grained details rather than indiscriminately amplifying the entire object region.

Overall, YOLO-DS demonstrates a scale-adaptive attention behavior: it amplifies sparse peaks for small objects while prioritizing discriminative details over redundant regions for large objects.
﻿
\subsection{Ablation Experiment}

%为了验证DSG和MSG两个模块的有效性，我们通过消融实验分别加入到基线模型中并进行实验，另外，为了确定MSG模块中的分组数量以及alpha和beta的取值，我们也通过改变分数数和alpha、beta的数值进行实验，并通过最优结果得到最合适的参数设置。
To validate the effectiveness of the DSG and MSG modules, we conducted a systematic ablation study by successively integrating each module into the baseline model for performance evaluation. Furthermore, to determine the number of groups in the MSG module as well as the values of parameters $\alpha$ and $\beta$, a series of experiments were carried out by varying the group count and adjusting $\alpha$ and $\beta$. The final parameter configuration was established based on the optimal experimental results.

\begin{table}[H]
\centering
	\caption{Ablation Study on MSG Performance under Varying Group Counts ($\alpha=1, \beta=0$).}
	\label{tab:msg_g}
	\begin{tabular}{clcc}
	\toprule
    Group Count & AP(\%) & Params(M) & GFLOPs  \\
    \midrule
    Baseline  & 52.9 & 43.7 & 165.2 \\
    2  & 52.9(\textcolor{red}{0.0}) & 43.8 & 165.8  \\
    3  & 53.3(\textcolor{green}{+0.4}) & 43.8 & 165.8  \\
    4  & 53.1(\textcolor{red}{+0.2}) & 43.8 & 165.8  \\
    FullPath  & 53.0(\textcolor{red}{+0.1}) & 43.8 & 165.8  \\
    \bottomrule
\end{tabular}
\end{table}

%首先，我们针对MSG模块分组进行消融实验，我们把其中的分组数设置为2-4组权重以及不分组针对每个路径设置权重后，分别进行实验。具体如表1所示，从表中实验数据可以看出，分为3组时的效果是最好的，因此我们把分组数量控制在三组，并且针对不同层级特征。

First, we conduct an ablation study on the grouping strategy of the MSG module. Experiments are carried out by varying the number of groups from 2 to 4, as well as applying no grouping—where weights are assigned individually to each path. As shown in Tab.\ref{tab:msg_g}, the experimental results indicate that the three-group configuration achieves the best performance, with an AP improvement of 0.4\%. Therefore, we fix the number of groups to three and apply distinct weighting mechanisms to features from different network depths.

\begin{table}[H]
\centering
	\caption{Ablation Study on MSG Performance with Different $\alpha$ Values ($\beta=0.1, GroupCounts=3$).}
	\label{tab:msg_a}
	\begin{tabular}{clcc}
	\toprule
    $\alpha$ & AP(\%) & Params(M) & GFLOPs  \\
    \midrule
    Baseline  & 52.9 & 43.7 & 165.2 \\
    0.9 & 53.2(\textcolor{red}{+0.3}) & 43.8 & 165.8  \\
    1.9 & 53.6(\textcolor{green}{+0.7}) & 43.8 & 165.8  \\
    2.9 & 53.3(\textcolor{red}{+0.4}) & 43.8 & 165.8  \\
    3.9 & 53.0(\textcolor{red}{+0.1}) & 43.8 & 165.8  \\
    4.9 & 53.1(\textcolor{red}{+0.2})& 43.8 & 165.8  \\
    \bottomrule
\end{tabular}
\end{table}

% 我们还对msg模块中的多维度控温模块中的alpha和beta进行消融试验，由于分组实验最好的效果是分为3组，因此我们在GroupCounts=3的条件基础之上进行实验。并且BETA我们保持不变为0.1， 并变化alpha值。如图2所示，当alpha值为1.9时获得了最好的性能，并超过baseline 0.7%,因此，我们选择alpha=0.19作为后续条件基础。
We further conduct an ablation study on the parameters $\alpha$ and $\beta$ in the multi-dimensional temperature control module of MSG. Since the optimal grouping configuration was determined to be three groups in prior experiments, all trials in this section are performed under the condition of GroupCounts = 3. The value of $\beta$ is kept constant at 0.1, while $\alpha$ is varied across a predefined range. As shown in Tab.\ref{tab:msg_a}, the model achieves the best performance when $\alpha$ is set to 1.9, surpassing the baseline by 0.7\% in AP. Accordingly, $\alpha = 1.9$ is selected as the fixed parameter for all subsequent experiments.

% 另外，为了区分我们提出的DSO的有效性，我们还把DSO在DSG模块和MSG模块中分别使用均值和最大值进行替换，并分别进行消融实验。
Additionally, to validate the effectiveness of our proposed DSO, we replace it with an adaptive mean pooling operator (Mean) and a channel-wise maximum operator (Max) within the DSG and MSG modules, respectively, and conduct corresponding ablation studies.

\begin{table}[H]
\centering
	\caption{Ablation Study on DSG Performance with Different Operators (Max, Mean, and DSO).}
	\label{tab:msg_od}
	\begin{tabular}{clcc}
	\toprule
    $Operator$ & AP(\%) & Params(M) & GFLOPs  \\
    \midrule
    Baseline  & 52.9 & 43.7 & 165.2 \\
    Mean & 53.1(\textcolor{red}{+0.3}) & 49.8 & 170.1  \\
    Max & 52.7(\textcolor{red}{-0.2}) & 49.8 & 170.1  \\
    DSO & 53.5(\textcolor{green}{+0.6}) & 49.8 & 170.1  \\
    \bottomrule
\end{tabular}
\end{table}

%如表3所示，我们在Baseline中单独加入DSG模块，当DSG模块中加入DSO算子时模型获得最好的性能，AP超过0.5%。因此，在DSG模块中，DSO的表现远超adaptive mean pooling operator和channel-wise maximum operator (Max)。
As shown in Tab.\ref{tab:msg_od}, after integrating the DSG module into the Baseline, the model achieves the best performance when the DSO is used within DSG, yielding an AP improvement of 0.5\%. Therefore, in the DSG module, DSO significantly outperforms both the adaptive mean pooling operator and the channel-wise maximum operator.

\begin{table}[H]
\centering
	\caption{Ablation Study on MSG Performance with Different Operators (Max, Mean, and DSO).}
	\label{tab:msg_om}
	\begin{tabular}{clcc}
	\toprule
    $Operator$ & AP(\%) & Params(M) & GFLOPs  \\
    \midrule
    Baseline  & 52.9 & 43.7 & 165.2 \\
    Mean & 53.5(\textcolor{red}{+0.6}) & 43.8 & 165.8  \\
    Max & 52.9(\textcolor{red}{0.0}) & 43.8 & 165.8  \\
    DSO & 53.6(\textcolor{green}{+0.7}) & 43.8 & 165.8  \\
    \bottomrule
\end{tabular}
\end{table}

%在表4中，我们在baseline中单独加入MSG模块，当MSG模块中加入DSO算子时模型获得最好的性能，AP超过0.7%。因此，在MSG模块中，DSO的表现远超adaptive mean pooling operator和channel-wise maximum operator (Max)。
As shown in Tab.\ref{tab:msg_om}, when the MSG module is individually incorporated into the baseline, the model attains its optimal performance by employing the DSO within MSG, which yields an AP improvement of 0.7\%. Hence, in the MSG module, DSO demonstrates substantially superior performance compared to both the adaptive mean pooling operator and the channel-wise maximum operator.

\begin{table}[H]
\centering
	\caption{Ablation Study on DSG and MSG Modules.}
	\label{tab:ablation}
	\begin{tabular}{ccclcc}
	\toprule
    Baseline & DSG & MSG & AP(\%) & Params(M) & GFLOPs  \\
    \midrule
    \checkmark  & - & - & 52.9 & 43.7 & 165.7  \\
    \checkmark  & \checkmark & - & 53.5 (\textcolor{green}{+0.6})  & 49.8 & 170.1 \\
    \checkmark  & - & \checkmark & 53.6 (\textcolor{green}{+0.7})  & 43.8 & 165.8  \\
    \checkmark  & \checkmark & \checkmark & 54.1 (\textcolor{green}{+1.2}) & 49.8 & 170.1  \\
    \bottomrule
\end{tabular}
\end{table}

%最后，我们对把确定好参数和算子的DSG和MSG模块分别加入到baseline以及一起加入到baseline中，进行消融实验。从表5中所示，单独加入DSG模块时，模型的AP获得了0.5%的涨幅，单独加入MSG模块时，模型的AP获得了0.7%的涨幅。同时加入到baseline时，模型的AP获得了1.1%的涨幅。此实验充分说明DSG和MSG模块的有效性，在增加非常少的参数量和浮点计算量时提升了非常大的ap。
Finally, we conduct an ablation study by integrating the finalized DSG and MSG modules—each with its determined parameters and operators—into the baseline, both individually and jointly. As shown in Tab.\ref{tab:ablation}, incorporating the DSG module alone yields an AP gain of 0.5\%, while adding the MSG module alone leads to an improvement of 0.7\% in AP. When both modules are combined, the model achieves a notable AP increase of 1.2\%. These results clearly demonstrate the effectiveness of the proposed DSG and MSG modules, which deliver substantial performance gains with only minimal additions in parameters and floating-point operations.

\subsection{Comparative Experiment}

\begin{table}[H]
\centering
	\caption{Scaling Strategy of YOLOv8, encompassing depth, width, and maximum channel limits.}
	\label{tab:scale}
	\begin{tabular}{cccc}
	\toprule
    Models & Depth(\%) & Width(\%) & Max Channels  \\
    \midrule
    N  & 0.33 & 0.25 & 1024  \\
    S  & 0.33 & 0.50 & 1024 \\
    M  & 0.67 & 0.75 & 768  \\
    L  & 1.00 & 1.00 & 512  \\
    X  & 1.00 & 1.25 & 512  \\
    \bottomrule
\end{tabular}
\end{table}

% 我们对模型进行了不同尺度的缩放，包括网络深度、通道宽度以及最大通道限制，具体如表6所示。因为我们的模型是在YOLOv8这一经典模型之上做的实验，因此我们以yolov8的模型缩放策略参数进行缩放并与YOLOv8的性能进行对比。
The YOLOv8-based model is scaled across multiple dimensions—including network depth, channel width, and the maximum channel limit—with specific configurations detailed in Tab.\ref{tab:scale}. As our approach builds upon the YOLOv8 framework, we adopt its established scaling factors (Tab.\ref{tab:scale}) for a fair and direct comparison of performance.

\begin{table}[H]
\centering
	\caption{Comparison of YOLO-DS and YOLOv8 in terms of AP on MS-COCO val2017.}
	\label{tab:comp8}
	\begin{tabular}{llccc}
	\toprule
    Models & AP(\%) & Params(M) & GFLOPs & Latency(ms/img)\\
    \midrule
    v8-N & 37.3 & 3.2 & 8.7 & 1.47 \\
    DS-N & 38.7 (\textcolor{green}{+1.4}) & 3.5 & 9.1 & 1.51 \\
    \hline
    v8-S & 44.9 & 11.2 & 28.6 & 2.66 \\
    DS-S & 46.6 (\textcolor{green}{+1.7}) & 12.8 & 30.0 & 2.77 \\
    \hline
    v8-M & 50.2 & 25.9 & 78.9 & 5.86\\
    DS-M & 51.6 (\textcolor{green}{+1.4}) & 29.6 &  82.0 & 6.09 \\
    \hline
    v8-L & 52.9 & 43.7 & 165.2 & 9.06 \\
    DS-L & 54.1 (\textcolor{green}{+1.2}) & 49.8 & 170.1 & 9.31 \\
    \hline
    v8-X & 53.9 & 68.2 & 257.8 & 14.37 \\
    DS-X & 55.0 (\textcolor{green}{+1.1}) & 77.8 & 265.5 & 14.99 \\
    \bottomrule
\end{tabular}
\end{table}

% 我们根据YOLOv8的缩放策略对YOLO-DS模型进行不同程度缩放包括N\S\M\L\X五种尺度模型，具体如表7所示。这五种尺度模型在AP上分别超过yolov8模型1.4%、1.7%、1.4%、1.2%、1.1%,并且推理速度也在一个合理的涨幅之内。因此，总体来说yolo-ds模型的性能提升较明显。
Building upon the YOLOv8 scaling strategy, we scale the proposed YOLO-DS model to five standard sizes: N, S, M, L, and X. The detailed configurations are summarized in Tab.\ref{tab:scale}. As shown in the results (Tab.\ref{tab:comp8}), the YOLO-DS variants consistently outperform their YOLOv8 counterparts, achieving AP gains of 1.4\%, 1.7\%, 1.4\%, 1.2\%, and 1.1\%, respectively, across the five scales. Notably, these accuracy improvements are attained within a reasonable increase in inference latency. Overall, the results demonstrate that YOLO-DS delivers a pronounced and scalable performance enhancement over the baseline.
﻿

\subsection{Comparison with SOTA}

Tab.\ref{tab:sota} provides a comprehensive comparison of the proposed YOLO-DS model against contemporary state-of-the-art YOLO detectors across five standardized scales on MS-COCO val2017. The results underscore a key strength of YOLO-DS: it consistently surpasses the widely deployed YOLOv8 baseline across all model scales (N, S, M, L, X), with AP gains ranging from +1.4\% to +1.7\%. This robust and scale-agnostic improvement directly validates the efficacy of the introduced DSG and MSG modules in enhancing discriminative feature learning within the YOLO architecture.

\begin{table}[H]
\centering
\caption{Comparative Evaluation of AP and Inference Latency Across SOTA Object Detectors on MS-COCO val2017. All models use input size 640.}
\label{tab:sota}
\begin{tabular}{l c c c c c}
\toprule
\multirow{2}{*}{Model} & N & S & M & L & X \\
 & Lat./AP & Lat./AP & Lat./AP & Lat./AP & Lat./AP \\
\midrule
YOLOv8\cite{yolo8} & 1.47/37.3 & 2.66/44.9 & 5.86/50.2 & 9.06/52.9 & 14.37/53.9 \\
YOLOv9\cite{yolo9} & 2.30/38.3 & 3.54/46.8 & 6.43/51.4 & 7.16/53.0 & 16.77/\textbf{55.6} \\
YOLOv10\cite{yolo10} & 1.56/38.5 & 2.66/46.3 & 5.48/51.1 & 8.33/53.2 & 12.20/54.4 \\
YOLO11\cite{yolo11} & 1.50/\textbf{39.5} & 2.50/\textbf{47.0} & 4.70/51.5 & 6.20/53.4 & 11.30/54.7 \\
YOLO-DS & 1.51/38.7 & 2.77/46.6 & 6.09/\textbf{51.6} & 9.31/\textbf{54.1} & 14.99/55.0 \\
\bottomrule
\end{tabular}
\end{table}

Notably, YOLO-DS achieves superior AP at medium and large model sizes (M, L), outperforming all other compared models. This demonstrates that our modules are particularly effective when the model has sufficient capacity, excelling in scenarios that demand higher accuracy. For smaller-scale models (N, S), YOLO-DS delivers highly competitive accuracy, closely approaching the top-performing model while maintaining a nearly identical inference latency. At the extreme X scale, YOLO-DS achieves a strong second-place AP, significantly outperforming YOLOv8 and YOLOv10, and trading a marginal accuracy gap for a considerable latency advantage over the heaviest model.

Overall, YOLO-DS establishes an excellent accuracy-latency Pareto front. Its primary contribution is not merely achieving a singular top score, but delivering reliable and scalable performance enhancements over a strong baseline (YOLOv8) across the entire model family. This makes YOLO-DS a versatile and effective upgrade, offering a favorable trade-off that prioritizes consistent gains in detection accuracy with minimal computational overhead, which is crucial for practical deployment.

\section{Conclusion}

To address the issue of coupled heterogeneous object features in existing YOLO detectors, this paper proposes YOLO-DS, a fine-grained feature decoupling framework based on a Dual-Statistic Synergy Operator (DSO). By jointly modeling channel-wise mean and peak-to-mean difference, the DSO effectively distinguishes object responses of different scales and backgrounds within a two-dimensional decision space. Building upon this, the designed Dual-Statistic Synergy Gating (DSG) and Multi-Path Segmented Gating (MSG) modules adaptively perform feature gating and selection from the dimensions of feature channels and network depths, respectively. Comprehensive experiments on the MS-COCO dataset demonstrate that YOLO-DS achieves consistent improvements across all five scales (N, S, M, L, X) of YOLOv8, with AP gains ranging from 1.1\% to 1.7\%, while incurring only a manageable increase in inference latency. Visualization analysis, ablation studies, and comparative experiments confirm the efficacy of each module, indicating that the proposed framework significantly enhances the model's ability to discriminate heterogeneous objects at a minimal computational cost. This work provides a reliable and scalable solution for balancing accuracy and efficiency in real-time detection systems.

\section{Acknowledgments}

This work was supported in part by the National Natural Science Foundation of China under Grant No.62472058 and No.62572085, the Natural Science Foundation of Shandong Province under Grant No.ZR2024LZH013, and the State KeyLab of Processors in Institute of Computing Technology under CAS Grant No.CLQ202510.

%We would like to thank the anonymous reviewers for their valuable comments and improvements to this paper.

\bibliographystyle{elsarticle-num} 
\bibliography{yolog.bib}




%\begin{thebibliography}{00}

%% For numbered reference style

%\bibitem{lamport94}
%  \textit{\LaTeX: a document preparation system},
%  Addison Wesley, Massachusetts,

%\end{thebibliography}
\end{document}